\documentclass[11pt,journal,compsoc,onecolumn]{IEEEtran}
\usepackage{cite}
\usepackage{times}
\usepackage{epsfig}
\usepackage[ruled,lined,linesnumbered,boxed]{algorithm2e}
\usepackage{color}
\usepackage{amsmath,graphicx,multirow}
\usepackage{bbm,amssymb}
\usepackage{dsfont}
\usepackage{algorithmic}

\def\K{{\bf K}}

\def\P{{\bf P}}
\def\t{{\bf \textrm{tr}}}
\def\Y{{\bf Y}} 
\def\x{{\bf x}} 
\def\S{{\cal S}}

\def\KK{{\cal K}}

\def\ds1{\mathds{1}}
\def\PPhi{{\bf \Phi}}

\begin{document}

\title{Deep Context-Aware Kernel Networks}
 
\author{Mingyuan Jiu, 
        Hichem Sahbi
 \thanks{M. Jiu is with School of Information Engineering, Zhengzhou University. Zhengzhou, 450001, China. Email: iemyjiu@zzu.edu.cn.} 
\thanks{H. Sahbi is with CNRS LIP6, UPMC Sorbonne University, Paris, 75005, France. Email: hichem.sahbi@sorbonne-universite.fr.}
}

\maketitle

\begin{abstract}
  Context plays a crucial role in visual recognition as it provides complementary clues for different learning tasks including image classification and annotation. As the performances of these tasks are currently reaching a plateau, any extra knowledge, including context, should be leveraged in order to seek significant leaps in these performances.  In the particular scenario of kernel machines, context-aware kernel design aims at learning positive semi-definite similarity functions which return high values not only when data share similar contents, but also similar structures (a.k.a contexts). However, the use of context in kernel design has not been fully explored; indeed, context in these solutions is handcrafted instead of being learned. \\
In this paper, we introduce a novel deep network architecture that learns context in kernel design. This architecture is fully determined by the solution of an objective function mixing a content term that captures the intrinsic similarity between data, a context criterion which models their structure and a regularization term that helps designing smooth kernel network representations. The solution of this objective function defines a particular deep network architecture whose parameters correspond to different variants of learned contexts including layerwise, stationary and classwise; larger values of these parameters correspond to the most influencing contextual relationships between data.  Extensive experiments conducted on the challenging ImageCLEF Photo Annotation and Corel5k benchmarks show that our deep context networks are highly effective for image classification and the learned contexts further enhance the performance of image annotation.
 \end{abstract}


\IEEEpeerreviewmaketitle

\section{Introduction} \label{sec:intro}

\IEEEPARstart{F}{}ollowing the rapid development of electronic devices and social medias, there is an exponential growth of image and video collections in the web and this makes their manual annotation and search completely out of reach. This rapid growth greatly motivates the need for automatic solutions that help analyzing and indexing these large collections \cite{HungTsaiCVPR2107,sahbijstars17,MartinsJVCIR2014,ArunIJMLC2017,sahbiicassp13a,ZhangIET2018,JinMTA2019,ZhangEAAI2019,sahbiigarss12a}.  Among these solutions, image annotation is a major challenge, which aims at describing contents of images with multiple semantic concepts (a.k.a keywords, categories or classes)~\cite{sahbiicip18,Bernard2003, sahbicassp11,Makadia2008, boujemaa2004visual}, for different use-cases  including image retrieval, human-computer interaction, autonomous driving, etc. Image annotation is challenging as concepts are usually diverse ranging from simple objects (persons, dogs, cars, etc.), to abstract  notions (clouds, cloudless, etc.), through highly interacting scene parts (handshake, fight, etc.); hence, learning {\it pure visual content models (without context) is clearly insufficient} \cite{sahbicvpr08a,sahbipr2012,sahbicbmi08}.\\
\indent Most of the existing image annotation techniques are based on machine learning. These methods build decision functions that learn the intricate relationships between images and their semantic concepts using variety of models including Support Vector Machines (SVMs)~\cite{sahbijmlr06,Goh2005,sahbiphd,Qi2007,sahbi2002coarse}, Nearest Neighbor classifiers~\cite{napoleon2010,Guillaumin2009, sahbiicip09,Verma2012,sahbiclef08}, deep networks\cite{Krizhevsky2012,deng2014deep,sahbiiccv17,girshick2014rich,Goodfellowetal2016,icassp2017b,srivastava2015training,szegedy2015going,russakovsky2015imagenet,tristan2017}, etc. and the  membership of images to different semantic concepts is decided by the scores of these models. Among the aforementioned machine learning techniques, SVMs are highly effective; their general principle consists in mapping nonlinearly separable data from input spaces into high (possibly infinite) dimensional spaces and finding  hyperplanes that separate these data while maximizing their margin. This mapping is achieved (either explicitly or implicitly) using particular similarity functions referred to as kernels~\cite{ShaweTaylor2004,lingsahbi2013}. The latter defined as symmetric positive semi-definite (p.s.d) functions should reserve high values only when data share similar semantics and vice-versa. Several kernels have been proposed in the literature including linear, polynomial and RBF as well as histogram intersection~\cite{barla2003histogram,lingsahbiicip2014,Maji2008,lingsahbieccv2014,sahbikpca06}. These functions can also be combined in order to learn more relevant similarities using multiple~\cite{Bach2004}, additive~\cite{Vedaldi2012} and deep kernels~\cite{Jiu2015,Jiu2016a,Jiutip2017}. Nonetheless, standard kernels and their combinations rely mainly on the visual content of images which is highly variable and insufficient to capture the semantics of images; hence, context should also be leveraged in order to further improve the discrimination power of the learned kernels. With context, kernels should reserve high values not only when images share similar content but also similar context; as shown through this paper, when context is learned (instead of being handcrafted), the accuracy of the resulting kernels is further improved.  

Given an image as a constellation of cells (e.g., grid of patches~\cite{thiemert2005applying}), with each cell being described with a feature vector (Bag-of-Word histograms, CNN-based, etc.), we design an elementary kernel that captures the similarity between the content of these cells together with their context (see also \cite{Belongie01shapematching, Hecvpr2004, Sahbi2015,Sahbi2011,Jiuprl2014}). Our design principle is based on the optimization of an objective function mixing (i) a content term that captures the visual similarity of cells, (ii) a context criterion which models their structure and (iii) a  regularization term that allows us to obtain a smooth kernel solution. Besides, this particular regularizer makes it possible to define an explicit kernel map through a parametric neural network\footnote{this makes our kernel map very suitable to handle large scale databases as the complexity scales linearly w.r.t the size of  training and test data while in standard kernel based methods, this complexity is at least quadratic.} whose architecture is defined by the kernel solution and whose parameters correspond to the learned context; hence, each layer in this network corresponds to a particular setting of context.  The learning of the network parameters is achieved using an ``end-to-end'' framework that minimizes the loss of a multi-class SVM trained on top of the deep kernel network.  Note that this formulation is different from~\cite{Sahbi2015,Sahbi2013icvs,Sahbi2011} as context in this related work is handcrafted while in our proposed method, it is learned (see Section~\ref{sec:unslearning}) in order to further enhance classification performances~\cite{Sahbi2015}. A part from the extensive experiments on ImageCLEF and Corel5k data, we further enhance the performances by making context classwise learned and this results into an extra gain in performances as corroborated through experiments.

\noindent Considering these issues, the main contributions of this paper include:

\begin{itemize}
\item A novel method that learns effective (context-aware) kernels using deep networks. While the early formulation in~\cite{Sahbi2011,Sahbi2015} relies on rigid (fixed) contexts, the ones proposed in this paper are learned instead of being handcrafted. This is achieved as a part of an ``end-to-end'' training framework which shows superior performances compared to handcrafted contexts.   
\item  The study  of different variants of contexts including {\it layerwise, stationary and classwise}. Layerwise contexts are learned with different parameters (through the layers of our deep-net) while stationary ones assume shared contextual relationships between all the layers and this reduces the risk of overfitting. Besides, classwise contexts are also investigated in order to build class-dependent parameters; this generates multiple branches of contexts each one dedicated to a particular category in image classification.
  
\item All these statements are corroborated through extensive experiments in image classification using the challenging ImageCLEF and Corel5k benchmarks. 
\end{itemize}

The rest of this paper is organized as follows: first, we review the related work about context learning in Section~\ref{sec:relatedwork}, and then we revisit our previous context-aware kernel design ~\cite{Sahbi2013icvs, Sahbi2015} in Section~\ref{sec:unslearning}. In Section~\ref{sec:deepconstruction}, we introduce our two main contributions: i) a deep network that learns explicit context-aware kernel maps, and  ii) different variants of context learning (including layerwise, stationary and classwise) which model context and further enhance the classification performances. In particular, classwise context learning  makes it possible to build specific contexts for different classes. In Section~\ref{sec:experiments}, we show the performance and comparison of our method on the ImageCLEF Photo Annotation and Corel5k benchmarks.  Finally, we conclude the paper in Section~\ref{concl} and we provide possible extensions for a future work.

\section{Related work} \label{sec:relatedwork}

Prior to detail our main contribution (in sections~III and IV), we discuss in this section the related work both in image annotation and context modeling.

\subsection{Image annotation}

State of the art in image annotation can be categorized into two major families of methods: discriminative and generative (see for instance ~\cite{YLiupr2007, ZhangDPR2012, Cheng2018, Bhagat2018}). The latter aim at modeling the joint distribution between observed (training) data and their ground truth and use maximum  a priori/posteriori to infer concepts on unseen data while the former seek to learn dependencies (or relationships) between images and their classes through decision functions\footnote{e.g. SVMs~\cite{Goh2005, Grangier2008}, decision trees~\cite{Wong2008}, artificial neural networks~\cite{Kuroda2002}, etc.} that map visual features into semantic concepts. Amongst the models used for image annotation, those based on SVMs are particularly successful; in these methods,  each concept is treated as an independent class and a binary SVM is trained to predict the membership of its underlying concept into a given test image~\cite{Goh2005, sahbiclef13,Cusano2004}. \\
\indent Different SVM-based solutions have been proposed in the literature in order to further enhance the performance of image annotation. SVM ranking is used in ~\cite{Grangier2008} to achieve image annotation where relevant concepts are ranked higher than irrelevant ones while semi-supervised Laplacian SVM is considered in~\cite{Jiu2016a} in order to propagate concepts  from labeled to unlabeled images. SVMMN~\cite{Liu2018} is also proposed in order to improve the accuracy of SVM-based image annotation; it seeks to learn maximum margin classifiers with a minimum number of samples by modeling smoothness both at the sample and the concept levels together with a correlation criterion across classes.  Other methods proceed differently by learning discriminative kernels that improve the performance of binary SVMs; for instance, shallow and deep multiple kernel learning~\cite{Jiutip2017} are proposed for image annotation in order to combine different standard kernels. Context-dependent kernels~\cite{Sahbi2011, Sahbi2011a} are also proposed for multi-class image annotation; a variant of this method in~\cite{Sahbi2015} makes it possible to learn explicit maps of these kernels while being highly efficient.  Our proposed method in this paper follows this line and allows us to design context-aware kernels using deep context networks; the latter learn the geometric relationships in images using deep parametric networks which generate the maps that preserve the inner product of the original deep context dependent kernels while being highly efficient and effective. \\
\noindent Note that with the resurgence of deep convolutional neural networks (CNNs)~\cite{LeCun98,HeZhangCVPR2016}, a further ``impressive'' progress has recently been observed in the aforementioned image classification methods. This gain comes essentially from the high accuracy of the learned CNN representations that capture the visual content of images better than the widely used handcrafted representations~\cite{ZhengPami2018}. The common aspect of these techniques consists in revisiting and rebuilding annotation methods by learning classifiers\footnote{These classifiers include SVMs, voting, K-nearest neighbors, self-defined Bayesian models, etc.} on top of CNN representations instead of handcrafted ones (see for instance~\cite{Murthyicmr2015,NiuTIP2019,MaMTA2019}). Our proposed framework is also built on top of deep representations but {\it seeks to further enhance the accuracy of image annotation by integrating and learning context in kernel networks}. 

\subsection{Context modeling}

Context, as a complementary clue, has attracted a lot of interest in different computer vision applications ranging from 3D scene understanding~\cite{ZhangBaiICCV2107}, to scene parsing~\cite{HungTsaiICCV2107}, through object and person re-identification~\cite{HungTsaiCVPR2107}, etc. Early work includes ``shape context''~\cite{BelongieMalik2002}  which models the spatial relationships between image primitives (mainly interest points) in order to design handcrafted shape representations. Later, the pyramid matching kernel, introduced in~\cite{Grauman2007}, models similarities between multi-layout feature representations and the context-aware keypoint extractor (CAKE), in \cite{MartinsJVCIR2014}, makes it possible to describe and retrieve keypoints which are representative within a certain image context. A priori knowledge of human part relationships have also been leveraged into CNNs in order to design a spatially-constrained deep learning framework~\cite{Jiuprl2014} that captures more discriminative features for part segmentation. Context-aware kernel and its deep map variants~\cite{Sahbi2011, Sahbi2015} are proposed in order to design kernels accounting for ``image-image'' relationships. In our proposed method, we also design context-aware representations by modeling not only content but also context. However, in contrast to these related methods, context in this paper is learned (as a part of kernel network design) instead of being handcrafted. \\ 
\indent In the particular scenario of image annotation, several other methods leveraging contextual information have been proposed in the literature. Authors in \cite{ArunIJMLC2017} consider undirected graphical models that jointly exploit low-level features and contextual information (as concept co-occurrences and spatial correlation statistics) to classify local image blocks into predefined concepts. Zhang et al.~\cite{ZhangIET2018} propose a region annotation framework that exploits the semantic correlation of segmented image regions; this method assigns each segmented region to one concept and learns the relationships between labels and region locations using \textcolor{black}{PSA}. A hybrid annotation approach based on visual attention mechanism and conditional random fields is proposed in ~\cite{JinMTA2019}  in order to pay more attention to the salient regions during the annotation process. A tri-relational graph-based method (including image and region as well as label graphs) is proposed for web image annotation~\cite{ZhangEAAI2019}.\\
\noindent As stated earlier, our contribution is different from these works as we consider context learned (as a part of deep kernel network design) while in all these related methods, context is handcrafted. From the methodological point of view, our work is rather related to Convolutional Kernel Networks (CKN)~\cite{Mairal2014} which explicitly learn kernel maps for gaussian functions using convolutional networks. However, our work is conceptually different from CKN: on the one hand, our approach is not restricted to gaussian kernels and is capable of learning a more general class of kernels. On the other hand, no context is considered in CKN while our method incorporates contexts explicitly in kernel design. 

\section{Context-aware kernel maps}\label{sec:unslearning}

\begin{figure}[tbp]
\centering
\includegraphics[width=0.2\linewidth]{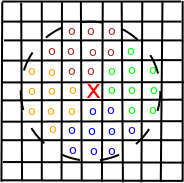}
\caption{This figure shows the handcrafted neighborhood system with four orientations to build the context-aware kernels. Red cross in the center means a particular cell in the regular grid, and colored circles around it within a radius of 3 stand for its 4 different sectors of neighbors (i.e. $C=4$).}
\label{fig:imagegrids}
\end{figure}

In this section, we briefly describe the design procedure of our context-aware kernel as well as its explicit map network. A context-aware kernel models the similarity between images using not only their content but also their context. The latter is relevant especially when the visual content of images  (belonging to the same semantic concepts) is noisy and highly variable. \\
\indent Considering $\{{\cal I}_p\}_p$ as a collection of images and ${\cal S}_p=\{ {\bf x}_1^p, \ldots, {\bf x}_n^p \}$ as a set of non-overlapping cells taken from a regular grid in ${\cal I}_p$ (see Fig.~\ref{fig:imagegrids}); without a loss of generality, $n$ is assumed constant for all images. We measure the similarity between any two images ${\cal I}_p$ and ${\cal I}_q$ using the convolution kernel defined as ${\cal K}( {\cal I}_p, {\cal I}_q) = \sum_{i, j} \kappa ({\bf x}_i^p, {\bf x}_j^q)$; here $\kappa$ is a p.s.d function (also referred to as elementary kernel) that defines the similarity between cells in ${\cal I}_p$ and ${\cal I}_q$. Resulting from the closure of the p.s.d with respect to the sum, the convolution kernel is also p.s.d. Note that ${\cal K}$ captures the similarity between two images ${\cal I}_p$ and ${\cal I}_q$ without necessarily aligning their cells, and this makes ${\cal K}$ translation and deformation resilient. In the above definition, one may consider $\kappa$ as one of the widely used standard kernels including linear, polynomial and gaussian as well as their multiple kernel combinations~\cite{Lanckriet2004}. However, these elementary kernels focus mainly on the visual content of primitives (cells) into images and ignore their contextual relationships. {\it A more relevant kernel $\kappa$ should provide high similarity values not only when primitives share close visual content but also similar context.} \\
\indent Following this goal, we propose to learn the gram matrix of $\kappa$ (denoted as  ${\mathbf{K}}$ with $\mathbf{K}_{\mathbf{x}, \mathbf{x}'}=\kappa({\mathbf{x}, \mathbf{x}'})$ and $\x$, $\x' \in {\cal X}$), by minimizing the following objective function~\cite{Sahbi2013icvs, Sahbi2015}
\begin{equation}
 \min_{\mathbf{K}} \t(-\mathbf{K}\mathbf{S}') - \alpha \sum_{c=1}^C \t (\mathbf{K} \mathbf{P}_c \mathbf{K}' \mathbf{P}_c^{'} ) + \frac{\beta}{2} ||\mathbf{K}||^2_2,
\label{equa:kernelfunction}
\end{equation}
\noindent here ${\cal X}=\cup_p {\cal S}_p$, $\beta > 0$, $\alpha \geq 0$, $'$ and $\t(.)$  stand for matrix transpose and the trace operator respectively. In the above objective function, $\mathbf{S}$ refers to a context-free kernel matrix between data in ${\cal X}$ (e.g., linear, RBF, etc.) and $\mathbf{P}_c$ refers to an intrinsic adjacency matrix that captures a particular spatial relationship between cells in $\cup_p {\cal S}_p$ (see more details subsequently). The left-hand side term of Eq.~(\ref{equa:kernelfunction}) is a fidelity criterion that provides high kernel values for visually similar pairs $\{(\x,\x')\}_{\x,\x'}$ while the second term makes the kernel values between these pairs stronger or weaker depending on the similarity of their neighbors. Finally, the right-hand side term acts as a regularizer that controls the smoothness of the learned kernel solution.\\
\noindent In our {\it initial} (handcrafted) definition  of context, we consider a {\it typed}  neighborhood system $\{\mathbf{P}_c\}_c$ instead of an isotropic (distance-based) one; the latter ---  in spite of being rotation invariant --- is known to be less discriminating~\cite{Sahbi2011, Sahbi2011a}. In contrast,  a typed system is more discriminating and can also be made invariant to different rigid transformations including rotations and scaling\footnote{One may estimate a ``characteristic'' orientation and scale of a given cell using the SIFT descriptor~\cite{lowe1999object}, and thereby make the typed adjacency matrices of the neighborhood system $\{\mathbf{P}_c\}_c$ rotation, scale and also translation invariant.}. In practice, we consider $C$ different adjacency matrices $\{\mathbf{P}_c\}_{c=1}^C$   (with $C=4$) corresponding to $4$ different relative positions of cells (namely ``above'', ``below'', ``left'' and ``right''; see also Fig.~\ref{fig:imagegrids}). Given a reference cell $\x$; {\it if $\x'$ is within a predefined range from $\x$ and with a relative position typed as $c$} then $\mathbf{P}_{c,\mathbf{x}, \mathbf{x}'}\leftarrow 1$; {\it otherwise} $\mathbf{P}_{c,\mathbf{x}, \mathbf{x}'}\leftarrow 0$. As shown later, this definition of context is {\it updated} by learning more relevant neighborhood systems (i.e., entries of $\{\mathbf{P}_c\}_{c=1}^C$)  that better fit our annotation tasks. \\

One may show that the optimization problem in Eq.~(\ref{equa:kernelfunction}) admits the following closed-form kernel solution
\begin{equation}
\mathbf{K}^{(t+1)} = \mathbf{S} + \gamma \sum_{c=1}^C \mathbf{P}_c \mathbf{K}^{(t)} \mathbf{P}_c^{'},
\label{equa:kernelsolution} 
\end{equation}
\noindent with $\mathbf{K}^{(0)}=\mathbf{S}$ and  $\gamma = \alpha / \beta$.  In this solution, $\gamma$ controls the influence of the context and it is chosen (in practice) in order to guarantee the convergence of the kernel solution to a fixed point. This property is guaranteed when $\gamma$ is upper bounded by the ratio of the norms of the right-hand side and the left-hand side terms of Eq.~(\ref{equa:kernelsolution}) (see more details in \cite{Sahbi2015}).\\
\indent Resulting from the p.s.d of $\mathbf{S}$ and the closure of the p.s.d with respect to the sum and the product, all the kernel matrices $\{\mathbf{K}^{(t)}\}_t$ defined in Eq.~(\ref{equa:kernelsolution}) are also p.s.d. Therefore, each kernel solution can be expressed as an inner product $\mathbf{K}^{(t)} =\mathbf{\Phi}^{(t)'} \mathbf{\Phi}^{(t)}$, with  $\mathbf{\Phi}^{(t)}$ being  an explicit feature map that takes data in $\cal X$ from an input space into a high dimensional Hilbert space. It follows that the solution of the objective function~(\ref{equa:kernelfunction}) can be equivalently rewritten as   
\begin{equation}
\mathbf{\Phi}^{(t+1)} = \Big( \mathbf{\Phi}^{'(0)} \  \   \gamma^{\frac{1}{2}} \mathbf{P}_1 \mathbf{\Phi}^{'(t)} \ \ldots \  \  \gamma^{\frac{1}{2}} \mathbf{P}_C \mathbf{\Phi}^{'(t)} \Big)'.
\label{equa:mapsolution}
\end{equation}

\noindent From Eq.~(\ref{equa:mapsolution}), it is clear that the dimensionality of $\mathbf{\Phi}^{(t)}$ is not constant and increases as $t$ evolves. However, when $\gamma$ is properly upper-bounded (see again~\cite{Sahbi2015}), the inner product $\mathbf{\Phi}^{'(t)}\mathbf{\Phi}^{(t)}$ is guaranteed to converge to a fixed-point provided that $T$ (with $t \leq T$) is sufficiently (but not very) large; as also observed in our experiments. Now considering the adjacency matrices $\{{\bf P}_c\}_c$ and a fixed (common) $T$ for all images, one may update the explicit kernel maps $\{\mathbf{\Phi}_\x^{(t)}\}_{t,\x}$ recursively ``image-by-image'' and this makes our kernel map evaluation not transductive.
\begin{figure}[tbp]
\centering
\scalebox{0.55}{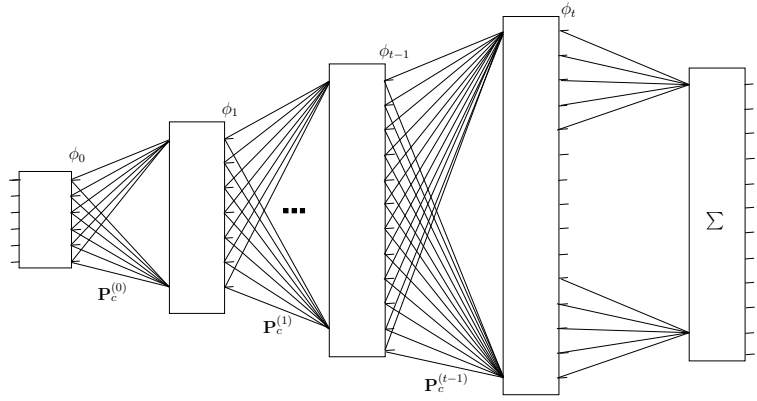}
\caption{This figure shows the ``unfolded'' multi-layered kernel map network with increasing dimensionality that captures larger and more influencing contexts. }\label{deep1}
\end{figure}
\begin{figure*}[htbp]
\begin{center}
\includegraphics[scale=0.35]{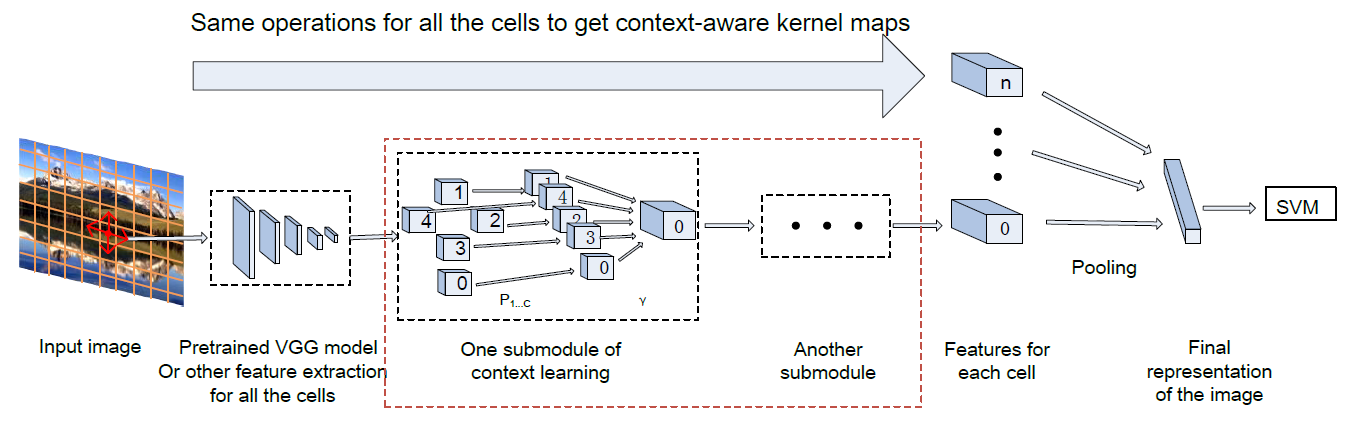}
\end{center}
\vspace{-0.5cm}
\caption{This figure shows the whole architecture and flowchart of our deep context learning.  Given an input image (divided into cells),  cells are first described using the pre-trained VGG-net. Afterwards, the context-based kernel map  of a given cell (for instance cell 0), at a given iteration, is obtained by combining the kernel maps of its neighboring cells (namely cells 1, 2, 3 and 4), obtained at the previous iteration, as shown in the red dashed rectangle and also in Eq.~(\ref{equa:mapsolution}). At the end of this iterative process,  the kernel maps of all the cells are pooled together in order to obtain the global representation of the input image, prior to achieve its classification. Note that  the network  shown in the red rectangle, together with the pooling layer, correspond   to the deep net shown in Fig. \ref{deep1}.} \label{fig:flowchart}
\end{figure*}
\section{Context Learning with Deep Networks} \label{sec:deepconstruction}
The method described in Section~\ref{sec:unslearning} is totally unsupervised as ground truth is not employed neither in kernel design nor in context definition. In order to further explore the potential of this method, we consider in this section, a supervised setting that allows us to learn more discriminating contexts as a part of kernel network design and this turns out to be more effective as shown later in experiments.
\subsection{From context-aware kernels to deep context networks} \label{sec:networkconstruction}
Given two cells $\x$, $\x'$ in $\cal X$ and following Eqs.~(\ref{equa:kernelsolution}) and (\ref{equa:mapsolution}), one may rewrite the kernel definition ${\K}_{\x,\x'}^{(t)}$ at iteration $t$  as 
\begin{equation} 
\begin{array}{l}
{\K}_{\x,\x'}^{(t)}=\phi_t(\phi_{t-1}(...\phi_1(\phi_0(\x)))). \phi_t(\phi_{t-1}(...\phi_1(\phi_0(\x')))), 
\end{array}
\end{equation} 
\noindent with $\phi_{t}(\x)=\PPhi_\x^{(t)}$. According to the definition of the convolution kernel ${\cal  K}$, the similarity between two images ${\cal I}_p$, ${\cal I}_q$ can be rewritten as
\begin{equation}
\begin{array}{ll}\label{sck}
 \displaystyle {\cal  K}( {\cal I}_p, {\cal I}_q)  & =  \displaystyle  \sum_{\x \in \S_p} \sum_{\x' \in \S_q} {\K}_{\x,\x'}^{(t)}    \\ 
                           & = \displaystyle \sum_{\x \in \S_p}  \phi_t(\phi_{t-1}(...\phi_1(\phi_0(\x))))  \\ 
                    &  \displaystyle \ \  \ \ \  \ \ \ \ \ . \sum_{\x' \in \S_q}  \phi_t(\phi_{t-1}(...\phi_1(\phi_0(\x')))).
\end{array}
\end{equation} 
\indent Eq.~(\ref{sck}) defines an inner product between two recursive kernel maps. Each one corresponds to a multi-layered neural network  (see Fig.~\ref{deep1}) whose layers deliver feature maps with increasing dimensionalities that correspond to larger and more influencing contexts; a final layer is added in order to pool these feature maps through all the cells of a given image. It is easy to see that the architecture in Fig.~\ref{deep1} is similar to the ones widely used in deep learning with some differences; on the one hand, as discussed earlier, the number units and the depth are respectively determined by (i) the dimensionality of the kernel maps $\{\PPhi_\x^{(t)}\}_t$ and (ii) the asymptotic behavior of our kernel solution; in other words, by the maximum number of iterations (again denoted as $T$) that guarantees the convergence of Eq.~(\ref{equa:kernelsolution}). In practice, we found that $T$-layers (with $T=5$) are enough in order to observe this convergence on all (training and test) images~\cite{Sahbi2015}. On the other hand, the parameters of this deep network correspond to the entries of the neighborhood system $\{{\bf P}_c\}_c$ and the largest parameters capture the most influencing contextual relationships.\\
\indent Considering the limit of Eq.~(\ref{equa:kernelsolution}) as ${\bf \tilde{K}}$ and the underlying map in Eq.~(\ref{equa:mapsolution}) as ${\bf \tilde{\Phi}}$, the convolution kernel $\cal K$ between two given images ${\cal I}_p$ and ${\cal I}_q$ can be written as 
\begin{equation}
{\cal K}({\cal I}_p,{\cal I}_q) = \langle \phi_{\cal K}({\cal S}_p), \phi_{\cal K}({\cal S}_q) \rangle, 
\end{equation}
with 
\begin{equation}
\phi_{\cal K}({\cal S}_p) = \sum_{\x \in {\cal S}_p}  {\bf \tilde{\Phi}}_{\x},  \label{equa:featpooling}
\end{equation}
\noindent so each constellation of cells in an image ${\cal I}_p$ can be represented by a deep explicit kernel map $\phi_{\cal K}({\cal S}_p)$. It is worth noticing that the maps in Eqs.~(\ref{equa:mapsolution}) and (\ref{equa:featpooling}) rely on the initial setting of $\{\PPhi^{(t)}\}_t$ (i.e., when $t=0$). The latter could be obtained {\it exactly} for some kernels (including polynomial and histogram intersection) or approximated for others (such as RBF) using kernel principal component analysis~\cite{Sahbi2013icvs}. \\ 
\indent In order to fully investigate the potential of the feature maps in Eq.~(\ref{equa:featpooling}), we consider in the subsequent section an ``end-to-end'' framework that learns the neighborhood system $\{{\bf P}_c\}_c$. The underlying context network is learned on top of another pre-trained CNN; as shown later in Section~\ref{sec:experiments}, this context learning process enhances further the performance of image annotation.

\subsection{Deep context learning} \label{sec:contextlearning}

In this section, we extend the approach described earlier in order to learn context.  Considering $N$ training images $\{{\cal I}_p\}_{p=1}^N$ belonging to $K$ different classes, we define $\Y_k^p$ ($k \in \{1,\dots,K\}$) as the class membership of a given image ${\cal I}_p$: here $\Y_k^p=+1$ iff ${\cal I}_p$ belongs to the class $k$ and $\Y_k^p=-1$  otherwise. For each class $k$, we train a binary SVM on top of the deep context network in order to decide about the membership of $k$ into test images. The objective function associated to these multi-class SVMs (shown subsequently) makes it possible to define an ``end-to-end'' framework that learns {\it not only} the SVM parameters but also the weights associated to the neighborhood system $\{{\bf P}_c\}_c$ (i.e., context).\\ 

\indent Considering the pairs of training images and their labels $\{(\phi_\KK({\cal S}_p),\Y_k^p)\}_p$, the loss associated to the multi-class SVMs is defined as
\begin{equation}\label{eq8}
\min_{\{w_k\}_k,\{\P_c\}_c}   \displaystyle   \sum_{k=1}^K \frac{1}{2} ||w_k||^2 + C_{k}  \sum_{p=1}^{N} \max\big(0, 1-\Y_k^p f_k({\cal S}_p)\big),  
\end{equation}  
\noindent with $C_k>0$, $f_k({\cal S}_p)=w_{k}' \phi_\KK({\cal S}_p)$ and $w_{k}$ the training parameters of $f_k$. The first term, of the above objective function, is an $\ell_2$ regularization that seeks to maximize the  margins of  $\{f_k\}_k$ while the second criterion corresponds to the hing loss. Eq.~(\ref{eq8}) makes it possible to learn at least two variants of contexts $\{{\bf P}_c\}_c$: global and classwise. In what follows, we first describe how to learn global contexts and then we update our scheme in section~\ref{classdependent} in order to make these contexts class-dependent. \\

\noindent As it is difficult to {\it jointly} optimize the two sets of parameters $\{w_k\}_k, \{\P_c\}_c$, we adopt an alternating optimization procedure: at each iteration, we fix $\{\P_c\}_{c}$ and we learn $\{w_k\}_k$ and then vice-versa. When fixing $\{w_k\}_k$, the parameters $\{\P_c\}_{c}$ are learned using backpropagation and gradient descent. Let $\mathds{1}_{\{\}}$ denote the indicator function; considering the kernel maps $\{\phi_\KK({\cal S}_p)\}_p$ of training images and the gradient of the loss (\ref{eq8}) ---  also denoted as $E$ ---  w.r.t  the output of the context network
\begin{equation}
\frac{\partial E}{\partial \phi_\KK} = - \sum_{p=1}^N \sum_{k=1}^K C_k \Y_k^p w_k \mathds{1}_{\{1-\Y_k^p w_{k}' \phi_\KK({\cal S}_p)\}},
\label{equa:gradientl1}
\end{equation}
we apply (i) the chain rule in order to backpropagate $\{\frac{\partial E}{\partial \P_c}\}_c$ and then (ii) gradient descent to update the neighborhood system $\{\P_c\}_{c}$. When fixing $\{\P_c\}_{c}$, the parameters $\{w_k\}_k$ of the primal form of $\{f_k\}_k$ are given by 
\begin{equation}
w_k = \sum_{p=1}^N \Y_k^p \alpha_k^p \phi_\KK({\cal S}_p),
\label{equa:dualprimalsolution}
\end{equation}
here $\{ \alpha_k^p \}$ correspond to the parameters of the dual form of Eq.~(\ref{eq8}) trained using LIBSVM~\cite{Chang2011}. Note that this iterative optimization procedure is performed till convergence, i.e., when the values of the two sets of parameters  $\{w_k\}_k, \{\P_c\}_c$ remain stable through iterations and this is observed (in practice) in less than 100 iterations.

\subsection{Layerwise vs. stationary contexts} \label{sec:sharedcontext}
In Section~\ref{sec:contextlearning},  context  matrices $\{\P_c\}_c$   are layerwise learned  and this increases the total number of training parameters and thereby the risk of overfitting. Actually, layerwise contexts are not totally independent; for instance, one may infer,   using transitive closure,  high-order contexts from low-order ones. Considering stationary context matrices  $\{\P_c\}_c$  through all the layers (written for short as $\P$), the underlying gradient  is obtained using the chain rule as \\ 
\begin{equation}
  \begin{array}{lll}
    \displaystyle     \frac{\partial E}{\partial \P} &=& \displaystyle\frac{\partial E}{\partial \phi_{\cal K}}  \frac{\partial \phi_{\cal K}}{\partial \P},
    \end{array}\label{eq:dscngradientR} 
\end{equation}
with the left-hand side term being defined in  Eq.~(\ref{equa:gradientl1}). For  stationary $\P$, the right-hand side term can be expanded by averaging (or equivalently summing) the gradients of all the instances of $\P$ through layers $t \in \{1,\dots,T\}$, and using again the chain rule,  as 
\begin{equation}
  \begin{array}{lll}
       \displaystyle        \frac{\partial \phi_{\cal K}}{\partial \P}                                                                                                &=& \displaystyle \sum_{t=1}^{T}   \frac{\partial \phi_{\cal K}}{\partial \phi_T}    \left(   \frac{\partial \phi_T}{\partial \phi_{T-1}} \dots \frac{\partial \phi_{t+1}}{\partial \phi_{t}}  \right) \frac{\partial \phi_t}{\partial \P}.
    \end{array}\label{eq:dscngradientP} 
\end{equation}
 
\noindent It is easy  to see that the general form of the ``between parentheses'' term in Eq.~\eqref{eq:dscngradientP} reduces to $1$ when $t=T$ and $\frac{\partial \phi_T}{\partial \phi_{T-1}}$ when $t=T-1$. Finally, context parameters are updated by gradient descent  using the shared gradients in Eqs.~\eqref{eq:dscngradientR} and~\eqref{eq:dscngradientP} for all the instances of $\P$  through all the layers; note that the initial values in $\P$ are also shared through all the layers so maintaining shared gradients, across  epochs,  guarantees stationary context at convergence.  The latter reduces the actual number of training parameters and thereby the risk of overfitting.  
 
\subsection{Global vs classwise contexts}\label{classdependent}

As shown  subsequently in experiments (see section~\ref{sec:experiments}), the impact of context learning is already well established with a {\it global} neighborhood system. However, this gain could further be enhanced if one considers instead a classwise context. Indeed, the rational resides in the fact that scenes may have different structures depending on their concepts. {\it Generally speaking, the notion of sharing intermediate representations in deep nets is highly valid when considering the shared intrinsic properties of the learned object categories, however, context is extrinsic, so making the latter class-dependent is complementary and rather more appropriate.} \\ 
In order to investigate the validity of this conjecture, we update the model slightly: as the objective function $E$ can be written as the sum of classwise losses, one may split its gradient into $K$ terms each one corresponding to a particular class. Then, using these $K$ gradients, we update the underlying neighborhood systems (now indexed by their classes and denoted as $\{\P_c^k\}_{c,k}$). This results into $K$ different context network maps\footnote{also indexed by their classes.} $\{\phi_\KK^k(.)\}_k$ used in order to evaluate the underlying SVMs. \\ 
\noindent Note that during the learning process, we adopt a warm-start in order to accelerate the convergence of our iterative algorithm. Indeed, the learned parameters $\{\P_c^k\}_{c,k}$ are initially set using the global learned context $\{\P_c\}_{c}$ and updated at each iteration in $K$-steps; each step includes (i) a backpropagation of $\{\frac{\partial E}{\partial \P_c^k}\}_{c}$ through the layers  of the $k^{th}$ deep context network using the chain rule and (ii) a gradient descent in order to update the underlying classwise neighborhood system $\{\P_c^k\}_c$. Once all these contexts learned (and fixed), the parameters $\{w_k\}_k$ are updated as shown earlier, in section~\ref{sec:contextlearning}, with the only difference that  maps used in Eq.~(\ref{equa:dualprimalsolution}) are now class-dependent. 

\section{Experimental Validation} \label{sec:experiments}

In this section, we evaluate the performance of our deep context-aware kernel network  --- using  different variants of contexts  --- on the challenging ImageCLEF and Corel5k image annotation benchmarks~\cite{Villegas2013,Duygulu2002}. Image annotation is a multi-task classification problem; given an image $\cal I$, the  goal is to assign a list of concepts (a.k.a. keywords) to $\cal I$ depending on the values of the underlying  classifiers. 

\begin{table*}[t]
	\centering
\resizebox{\textwidth}{!}{	
	\begin{tabular}{|c|c|c|ccc|ccc|}
	\hline
	\multirow{2}{*}{Cells} & \multirow{2}{*}{$r$} & \multirow{2}{*}{Method} & \multicolumn{3}{c|}{BoW features} & \multicolumn{3}{c|}{VGG features} \\
        \cline{4-9}
	& & & LIN & POLY & HI & LIN & POLY & HI \\
	\hline
	\multirow{1}{*}{1$\times$1} & - & CF & 40.24/24.21/49.31 & 43.76/27.26/52.27 & 44.16/25.76/53.78 & 52.84/42.30/66.75 & 55.70/44.69/71.02 & 54.13/43.66/70.52 \\	
	\hline
	\multirow{4}{*}{2$\times$2} & - & CF & 39.55/24.36/47.40 & 41.61/24.73/50.27 & 44.61/27.30/53.71 & 53.82/42.32/68.04 & 56.82/44.41/70.66 & 55.15/43.69/69.85 \\
	\cline{2-3}
	& \multirow{2}{*}{1} & Handcrafted CA & 42.23/25.23/51.24 & 42.93/26.13/52.33 & 45.93/28.39/54.74 & 54.47/43.24/69.56 & 56.53/45.23/71.43 & 55.18/42.97/70.43  \\
                               & & L-Global CA & 43.50/26.05/\textbf{51.68} & 43.48/26.41/\textbf{52.93} & 46.10/27.22/55.15 & 56.01/45.26/\textbf{70.49} & 58.29/46.76/72.06 & 58.26/44.99/71.54 \\
                               &  & S-Global CA & 43.59/26.23/51.69 & 43.78/26.73/52.94 & 46.51/28.35/54.75 & 56.07/45.32/70.45 & 58.58/46.89/72.15 & 58.43/45.48/71.64 \\                               
                               & & L-Classwise CA & \textbf{44.34}/\textbf{27.78}/51.52 & \textbf{45.17}/\textbf{28.21}/52.91 & \textbf{46.22}/\textbf{27.31}/\textbf{55.16} & \textbf{57.36}/\textbf{47.36}/70.35 & \textbf{58.87}/\textbf{47.98}/\textbf{72.08} & \textbf{58.88}/\textbf{46.51}/\textbf{71.54}  \\
                                        \hline
          \hline
	\multirow{7}{*}{4$\times$5} & - & CF & 40.37/24.89/48.01 & 42.27/26.56/50.12 & 43.39/28.53/52.40 & 51.24/38.34/63.21 & 52.55/39.96/64.71 & 51.33/37.96/63.64 \\
	\cline{2-3}
	& \multirow{3}{*}{1} & Handcrafted CA & 41.67/26.05/50.95 & 42.88/26.40/51.60 & 44.94/29.22/53.50 & 51.71/38.97/63.98 & 53.09/39.66/65.01 & 51.61/38.94/64.15  \\
                               & & L-Global CA & 44.74/26.85/\textbf{52.86} & 44.39/26.04/\textbf{52.82} & 45.77/27.02/\textbf{54.74} & 53.55/41.09/\textbf{65.36} & 54.80/41.98/66.11 & 53.26/39.11/\textbf{65.37} \\
                               & & S-Global CA & 43.81/27.30/52.14 & 44.63/26.66/52.74 & 46.09/28.34/54.60 & 53.57/41.43/65.43 & 54.63/42.32/66.22 & 53.70/40.85/65.52 \\                               
                               & & L-Classwise CA & \textbf{46.16}/\textbf{28.04}/52.48 & \textbf{46.03}/\textbf{27.49}/52.70 & \textbf{47.59}/\textbf{28.63}/54.05 & \textbf{55.96}/\textbf{44.25}/65.33 & \textbf{56.66}/\textbf{44.13}/\textbf{66.50} & \textbf{55.50}/\textbf{41.33}/64.96  \\

	\cline{2-3}
	& \multirow{3}{*}{3} & Handcrafted CA & 41.88/26.26/50.97 & 43.31/26.98/51.78 &  44.78/29.42/53.78 & 51.89/39.35/64.45 & 52.85/39.84/65.11 & 51.70/38.36/64.36 \\
                               & & L-Global CA & 43.81/27.97/\textbf{51.80} & 44.20/25.91/52.51 & 45.81/28.65/\textbf{54.91} & 54.55/40.67/\textbf{65.35} & 55.01/42.56/66.45 & 53.65/39.51/\textbf{65.44} \\
                               & & S-Global CA  & 44.00/28.02/51.81 & 44.59/27.31/52.63 & 46.09/28.72/54.72 & 54.14/40.66/65.27 & 55.16/42.26/66.30 & 53.89/40.12/65.37\\                               
                               & & L-Classwise CA & \textbf{44.90}/\textbf{28.78}/51.72 & \textbf{45.35}/\textbf{27.31}/\textbf{52.60} & \textbf{47.65}/\textbf{30.80}/54.75 & \textbf{55.63}/\textbf{42.84}/65.34 & \textbf{56.41}/\textbf{44.87}/\textbf{66.67} & \textbf{55.24}/\textbf{42.51}/65.31 \\
    \cline{2-3}
          \hline
          \hline
	\multirow{9}{*}{8$\times$10} & - & CF & 39.70/24.41/46.60 & 41.69/26.01/49.30 & 41.32/25.05/49.51 & 45.34/30.82/56.43 & 47.07/31.98/58.20 & 45.45/30.14/57.91  \\
	\cline{2-3}
	& \multirow{3}{*}{1} & Handcrafted  CA & 40.62/24.62/48.32 & 42.61/26.29/50.86 & 42.63/26.28/50.45 & 45.80/31.19/57.57 & 46.73/31.37/58.50 & 46.36/30.65/58.47  \\
	& & L-Global CA  & 42.67/26.42/50.52 & 44.15/26.74/\textbf{52.15} & 45.20/26.41/\textbf{53.87} & 47.48/32.74/58.66 & 48.36/32.80/\textbf{59.73} & 48.80/32.66/\textbf{59.92} \\
                               & & S-Global CA  & 43.64/26.35/51.35 & 43.80/27.12/51.84 & 45.23/26.45/53.41 & 47.25/33.01/58.36 & 49.25/34.41/59.74 & 48.97/32.51/59.48\\                               
                               & & L-Classwise CA & \textbf{46.23}/\textbf{29.98}/\textbf{51.22} & \textbf{46.13}/\textbf{29.24}/52.08 & \textbf{46.72}/\textbf{28.38}/53.48 & \textbf{49.79}/\textbf{36.20}/\textbf{58.67} & \textbf{50.76}/\textbf{34.56}/59.34 & \textbf{50.98}/\textbf{34.55}/59.28  \\
	\cline{2-3}
	& \multirow{3}{*}{3} & Handcrafted CA & 40.74/25.28/48.51 & 42.61/26.64/51.11 & 42.70/26.38/50.84 & 46.25/31.70/57.74 & 47.24/31.92/59.46 & 46.78/30.92/58.63  \\
	& & L-Global CA & 44.22/26.43/\textbf{52.02} & 44.06/26.45/52.38 & 45.30/26.73/53.83 & 48.39/33.89/\textbf{58.97} & 48.98/34.02/60.46 & 48.81/32.85/\textbf{60.07} \\
                               &  & S-Global CA& 43.42/26.52/50.76 & 44.38/26.56/52.31 & 44.32/26.18/52.83 & 47.64/33.25/58.79 & 48.40/34.38/59.65 & 48.32/33.37/59.48 \\
                               & & L-Classwise CA & \textbf{45.50}/\textbf{28.71}/51.99 & \textbf{46.15}/\textbf{28.09}/\textbf{52.76} & \textbf{46.82}/\textbf{27.91}/\textbf{53.93} & \textbf{49.76}/\textbf{35.89}/58.47 & \textbf{51.16}/\textbf{37.22}/\textbf{60.47} & \textbf{50.41}/\textbf{35.24}/59.71  \\
    \cline{2-3}
	& \multirow{3}{*}{5} & Handcrafted CA & 41.01/25.31/48.88 & 43.20/26.52/51.49 & 42.93/26.68/51.28 & 46.76/31.84/57.90 & 46.49/31.77/59.10 & 46.88/31.09/58.69  \\
	& & L-Global CA & 44.00/26.56/51.96 & 44.95/26.01/\textbf{53.41} & 45.56/26.18/\textbf{53.98} & 47.86/33.21/\textbf{58.84} & 48.66/33.41/\textbf{60.44} & 48.42/32.74/59.54 \\
                               &  & S-Global CA  & 44.11/26.90/51.73 & 44.71/25.67/53.24 & 45.33/26.49/53.39 & 47.62/33.45/58.51 & 48.21/33.64/59.76 & 48.32/33.07/58.91 \\	
                               & &L-Classwise CA & \textbf{45.58}/\textbf{28.44}/\textbf{52.11} & \textbf{45.59}/\textbf{26.71}/53.32 & \textbf{46.30}/\textbf{27.37}/53.82 & \textbf{49.33}/\textbf{35.29}/58.67 & \textbf{50.35}/\textbf{35.67}/60.41 & \textbf{49.86}/\textbf{34.60}/\textbf{59.54}  \\
	\hline
	\end{tabular} 
	}
	\vspace{0.1cm}
	\caption{The performance (in $\%$) of different variants of context-aware kernel networks on ImageCLEF. The triple $\cdot/\cdot/\cdot$ stands for MF-S/MF-C/mAP. In these experiments, $r$ corresponds to the radius of the disk that supports context. ``L-Global'', ``S-Global'' and  ``L-Classwise'' stand respectively for ``Layerwise Global'', ``Stationary Global'' and ``Layerwise Classwise''  contexts.\label{tab:imageclefresults}}
\end{table*} 

\subsection{ImageCLEF benchmark} \label{sec:exper:imageclef}

This dataset consists of more than 250k images --- belonging to 95 categories (or concepts) ---  split into training, dev and test data; note that these concepts are not exclusive, so each image may belong to one or multiple categories.  In our experiments, we only consider the dev set\footnote{This set includes 1,000 images equally split between training and testing.} as the ground-truth is released on this subset only and images belonging to this subset are partitioned into  regular grids of $W \times H$ cells\footnote{as shown later, different granularities are considered for $W$ and $H$.}. Each cell is encoded using two types of features: {\it handcrafted} and {\it learned}; the former include the bag-of-word (BoW) histogram (evaluated on dense SIFT features and a code-book of 500  words) while the latter include deep VGG-net features. More precisely, the used VGG-net corresponds to ``imagenet-vgg-m-1024''~\cite{Chatfield14}; this model is pretrained on ImageNet and consists in five convolutional and three fully-connected layers. The outputs of the second fully-connected layer are used to describe the content of the cells in the regular grids. Note that, prior to describe all these training and test images, we rescale them to their median dimension of $400\times500$ pixels. 

In order to learn the context-aware kernel networks, we consider four different types of geometric relationships (see Fig.~\ref{fig:imagegrids})  and different  context-free kernel maps including linear, polynomial and histogram intersection (HI). The explicit maps of linear and polynomial kernels can be exactly obtained using identity and tensor product respectively while for histogram intersection these maps can be approximated using decimal-to-unary projections (see for instance~\cite{Sahbi2015} for more details). Using the above setting, we learn maximum margin classifiers on top of the deep context-aware kernel networks and we measure their performances using the F-scores (defined as harmonic means of recall and precision) both at the sample and concept levels (denoted respectively as MF-S and MF-C) as well as the mean average precision (mAP): higher values of these measures imply better performances. \\ 

\begin{table}[htb]
	\centering	
\resizebox{0.45\textwidth}{!}{	
	\begin{tabular}{c|ccccc}
	\hline
    & 2-layer-net & 3-layer-net & 4-layer-net& 5-layer-net \\
	\hline
     RE & 23.97 & 4.08 & 0.99 & 0.25  \\
	\hline
	\end{tabular}
	}
    \vspace{0.1cm}
    \caption{This table shows the relative errors (in \%) of different layers in the deep context-aware kernel networks. These performances correspond to ``Layerwise Global'' (L-Global) learned contexts.\label{tab:imageclef:error:layers}}
\end{table}

\begin{table}[htb]
	\centering
\resizebox{0.45\textwidth}{!}{	
	\begin{tabular}{c|ccccc}
	\hline
      	   & & MF-S & MF-C & mAP & runtime \\
	\hline
 \multirow{2}{*}{2-layer} & Handcrafted CA & 40.93 & 25.20 & 48.78 & - \\
           & L-Global CA & 42.64 & 25.04 & 50.98 & 45.31 \\
    \hline
 \multirow{2}{*}{3-layer} & Handcrafted CA & 40.74 & 25.28 & 48.51 & - \\
           & L-Global CA & 44.22 & 26.43 & 52.02 & 235.10 \\
    \hline           
 \multirow{2}{*}{4-layer} & Handcrafted CA & 41.03 & 25.55 & 48.99 & - \\
           & L-Global CA & 44.11 & 26.57 & 52.43 & 1021.71 \\
    \hline           
 \multirow{2}{*}{5-layer} & Handcrafted CA & 40.93 & 25.54 & 48.68 & - \\
           & L-Global CA & 44.27 & 25.12 & 52.15 & 7729.90 \\
	\hline
	\end{tabular}
	}
    \vspace{0.1cm}
    \caption{This table shows the performance of handcrafted vs. learned deep context-aware kernel networks; we compare handcrafted vs. layerwise global (L-Global) contexts on ImageCLEF. The MF-S/MF-C/mAP  measures are provided (in \%) and runtime performances (in \textit{s}) for one ``forward-backward'' iteration of backpropagation. \label{tab:imageclef:perf:layers}}
\end{table}
\noindent \textbf{Impact of network depth.} As already discussed, the depth of the network is defined by the number of iterations $T$ in Eqs.~(\ref{equa:kernelsolution}) and (\ref{equa:mapsolution}); larger values of $T$ imply deeper and convergent networks\footnote{under a particular setting of $\gamma$ (and when $T$ is sufficiently large), the inner products defined by the feature maps of the network are convergent.} but the underlying feature maps become high dimensional. In contrast, low values of $T$ make the network relatively shallow and compact but not convergent. Hence, the appropriate setting of the network depth (i.e., $T$) should tradeoff {\it convergence} and {\it compacity}; this also impacts the number of training parameters (thereby generalization) and the computational efficiency of the resulting network. \\
\noindent In practice, we measure the convergence of the network, between two consecutive layers, using the following relative error (RE) criterion
\begin{equation}
 \text{RE}^{(t)} =  \frac{1}{\vert \S_p \vert \vert \S_q \vert}\sum_{\x \in \S_p} \sum_{\x' \in \S_q} \frac{\vert \mathbf{K}^{(t)}_{\x, \x'}-\mathbf{K}^{(t-1)}_{\x, \x'} \vert}{\vert \mathbf{K}^{(t)}_{\x, \x'}+\mathbf{K}^{(t-1)}_{\x, \x'} \vert},
\end{equation}
this measure is evaluated using the initial normalized  neighborhood system, i.e., $\{\P_{c, \x, \x'}\slash \sum_{\x'' \in \mathcal{N}_c(\x)} \P_{c, \x, \x''}\}_c$ with $\mathcal{N}_c(\x)$ being a subset of neighbors of $\x$. Tab.~\ref{tab:imageclef:error:layers} shows this error as the network becomes deeper; it is clear that descent convergence is obtained for reasonably (not very) deep networks. Tab.~\ref{tab:imageclef:perf:layers} shows the underlying performances with handcrafted and learned contexts. From these results, we observe that when context is handcrafted, the impact of the depth is marginal while for learned context this impact is noticeable. Runtime performances are also reported and correspond to the cost of one ``forward-backward'' iteration during backpropagation; these performances are obtained on a workstation with 4 Intel-Xeon CPUs of 3.2GHz and 64G memory. It is clear that the learning process becomes cumbersome as the network gets deeper; hence, in order to make the learning reasonably tractable, we consider in the remainder of this paper a network architecture with three layers. \\

\noindent \textbf{Impact of the neighborhood system.} In order to model different context granularities, we consider multiple instances of regular grids (with $W \times H$ in $\{2 \times 2,4 \times 5,8 \times10\}$) and different settings of the  radius $r$ in the neighborhood system. It is clear that $r$ is constrained by $W$ and $H$; for instance, when the latter are equal to $1$, the radius $r$ cannot exceed $1$ so this particular configuration (shown in the first row of Tab.~\ref{tab:imageclefresults}) corresponds  to a holistic ({\it one-cell-grid}) context-free (CF) network. From Tab.~\ref{tab:imageclefresults}, we first observe a global negative impact of finer grids on the performances of CF networks on both BoW and VGG features. Indeed, finer cells --- deprived from context --- are not sufficiently discriminating and hence powerless to capture the semantic of images.  However, context-aware (CA) networks, even-though applied to finer cells, allow us to recover and enhance the discrimination power of these cells and also to substantially overtake the original CF performances by a significant margin, especially when context is learned; this gain is consistently {\it the highest through all the original kernel maps}, particularly when cells are encoded with VGG and when $H \times W=2 \times 2$ (with $r$ being necessarily set to $1$). For larger $H \times W$, we observe a moderate (and sometimes a negative) impact of larger $r$ on performances. We conjecture that, as the radius gets larger, cells in the neighborhood system $\{{\cal N}_c(.)\}_c$ are (more and more) densely connected; as a result, it becomes more difficult to learn relevant context from a huge combinatorial set of possible relationships in  $\{{\cal N}_c(.)\}_c$. \\  

\begin{table}[tb]
	\centering
\resizebox{0.4\textwidth}{!}{	
	\begin{tabular}{c|ccc}
	\hline
	Kernel & MF-S & MF-C & mAP \\
	\hline
    \hline
    GMKL(\cite{Varma2009})  & 41.3 & 24.3 & 49.1 \\
    2LMKL(\cite{Zhuang2011a}) & 45.0 & 25.8 & 54.0 \\
    LDMKL (\cite{Jiutip2017}) & 47.8 & 30.0 & 58.6  \\
    Handcrafted CA (\cite{Sahbi2015}) & 56.5 & 45.2 & 71.4 \\
    L-Global CA (proposed) & 58.3 & 46.8 & 72.1 \\
    S-Global  CA (proposed) & 58.6 & 46.9 & 72.2\\
    L-Classwise CA (proposed) & 58.9 & 48.0 & 72.1\\
	\hline
	\end{tabular}
	}
    \vspace{0.1cm}
    \caption{This table shows comparison of performances (in \%) between different kernel-learning  methods on ImageCLEF. The best results, on both global and classwise contexts, are obtained using polynomial kernel and VGG features on a grid of $2\times2$ cells with $r=1$. \label{tab:imageclef:comparsionresults}}
\end{table}
\begin{table}[tb]
	\centering
\resizebox{0.5\textwidth}{!}{	
	\begin{tabular}{c|ccc}
	\hline
    Comparison & MF-S & MF-C & mAP \\
	\hline
    \hline
     L-Global CA vs. Handcrafted CA & 100 & 75.0 & 100 \\
     S-Global CA  vs.  L-Global CA & 63.9 & 86.1 & 33.3 \\
     L-Classwise CA vs. L-Global CA & 100 & 100 & 36.1 \\
     L-Classwise CA vs. S-Global CA   & 97.2 & 97.2 & 55.6 \\
	\hline
	\end{tabular}
	}
    \vspace{0.1cm}
    \caption{\textcolor{black}{This table shows the statistical dependencies  (in \%) between different variants of learned and handcrafted contexts. For each comparison (``A vs. B''), we measure the percentage of times ``A is not worse than B'' in a pool of 36 runs (using evaluation measures of Tab.~\ref{tab:imageclefresults}).}  \label{tab:imageclef:paircomparison}}
\end{table}
\noindent \textbf{Impact of stationary and classwise contexts.} In the following experiments,  the initial weights of classwise and stationary context networks are taken from the learned global context. Again, Tab.~\ref{tab:imageclefresults} shows the results of different context variants using several kernel map initializations and features, while Tab.~\ref{tab:imageclef:paircomparison} shows the statistical dependencies between different network pairs. From all these results, we observe that the learned context networks overtake the handcrafted ones over all the settings in MF-S/mAP and $75\%$ of the settings in MF-C. We also observe that stationary context networks are able to further enhance the performances of the global context networks for most of the settings, however the gain in mAP is less marked. In the subsequent results, the learned classwise context networks boost further the performances for most of the settings compared to the two other variants; for instance, MF-S/MF-C/mAP values raise from $42.67$/$26.42$/$50.52$ to $46.23$/$29.98$/$51.22$ (using linear kernel map with BoW features, $r=1$ and $W\times H=8\times10$)  and from $55.01$/$42.56$/$66.45$ to $56.41$/$44.87$/$60.67$ (using polynomial kernel map with VGG features, $r=3$ and $W\times H=4\times5$). In sum, {\it learned classwise context networks are more positively influencing than learned global ones, whether context is stationary or not} through different layers. \\ 

\noindent \textcolor{black}{\textbf{Qualitative results.} In order to visually analyze the learned contexts, we accumulate and display the weights involved in $\{\mathbf{P}_c\}_c$ (following the spatial support shown in Fig.~\ref{fig:imagegrids}).  We investigate two aspects: the interpretation of context evolution through layers and the interpretation of different context variants. Fig.~\ref{fig:contextevolution} (second, third and fourth rows) respectively describe the handcrafted and the learned  $\{\mathbf{P}_c\}_c$ in the first and the second layers of the underlying network; values in $\{\mathbf{P}_c\}_c$ are superimposed on two images of ImageCLEF. This display shows that {\it first} layer context is less meaningful than the {\it second} layer one, possibly resulting from the fact that the {\it latter} captures higher-order and more influencing spatial and structural relationships compared to the {\it former} which relies only on immediate neighbors in $\{\mathbf{P}_c\}_c$. Besides, due to the chain rule, the gradient w.r.t. the first layer context is more quickly vanishing and this makes its evolution through the iterations of back-propagation relatively less important compared to the gradient in the second layer.  Fig.~\ref{fig:contextexamples} (second and third rows) is a visualization of handcrafted vs. learned contexts (including layerwise, stationary and classwise) taken from the second layer of their respective networks superimposed on two images from ImageCLEF; it is clear that when contexts are learned, some spatial cell-relationships are {\it amplified} while others are {\it attenuated} and this reflects their importance in the underlying image classification tasks.} \\

\noindent {\textbf{Further comparison.} Finally, we compare the performance of the proposed approach to other methods on the ImageCLEF dataset. This comparison involves the most related kernel design techniques including: general multiple kernel learning (GMKL), two-layer multiple kernel learning (2LMKL) and Laplacian-based semi-supervised learning on 3-layer kernel network (LDMKL). Comparative results are shown in Tab.~\ref{tab:imageclef:comparsionresults}. The proposed classwise context-aware kernel networks obtain the best performance. The first row of Fig.~\ref{fig:annotationexamples} shows some image instances and their annotation results respectively using context-free, handcrafted and learned (layerwise vs. stationary and global vs. classwise) context networks.

\begin{table*}[ht]
	\centering
\resizebox{\textwidth}{!}{	
	\begin{tabular}{|c|c|c|ccc|ccc|}
	\hline
	\multirow{2}{*}{Cells} & \multirow{2}{*}{$r$} & \multirow{2}{*}{Method} & \multicolumn{3}{c|}{BoW features} & \multicolumn{3}{c|}{VGG features} \\
        \cline{4-9}
	& & & LIN  & POLY & HI & LIN & POLY & HI \\
	\hline
	\multirow{1}{*}{1$\times$1} & - & CF & 22.99/15.06/13.05/122 & 26.90/18.47/16.26/147 & 22.69/18.62/14.38/131 & 42.26/29.10/28.15/179 & 46.91/31.04/30.15/191 &  46.49/30.70/29.41/195 \\
	\hline
	\multirow{4}{*}{2$\times$2} & - & CF & 23.36/16.21/13.75/128 & \textbf{26.12}/17.98/14.89/137 & 22.76/16.94/13.33/125 & \textbf{44.75}/35.25/31.88/190 & 43.96/34.91/31.49/188 & 42.88/33.77/29.42/186  \\
	\cline{2-3}
	& \multirow{2}{*}{1}  & Handcrafted CA & 24.66/17.53/14.60/129 & 25.68/18.46/15.17/137 & 24.66/17.95/15.04/131 & 43.48/35.71/32.07/189 & 43.70/34.95/31.56/186 & \textbf{44.15}/33.85/30.03/188  \\
                               & & L-Global CA & 24.88/17.77/14.83/135 & 25.37/18.77/15.20/137 & 24.51/19.26/15.73/131 & 43.19/35.73/31.94/189 & 44.65/35.76/32.40/189 & 42.73/33.89/29.64/186  \\
          	 & & S-Global CA & 26.29/18.74/15.88/138 & 26.10/19.53/16.28/141 & 25.58/19.89/16.52/138 & 44.70/36.66/33.07/193 & 45.50/36.32/32.78/194 & 45.28/34.63/31.02/194  \\
                               & & L-Classwise CA & \textbf{26.42}/\textbf{18.45}/\textbf{15.75}/\textbf{138} & 25.35/\textbf{19.25}/\textbf{15.76}/\textbf{138} & \textbf{26.11}/\textbf{20.42}/\textbf{16.71}/\textbf{142} & 43.74/\textbf{36.41}/\textbf{32.59}/\textbf{190} & \textbf{45.31}/\textbf{36.36}/\textbf{32.76}/\textbf{192} & 43.51/\textbf{34.93}/\textbf{30.37}/\textbf{188} \\
          \hline
          \hline
	\multirow{7}{*}{4$\times$5} & - & CF & 21.71/15.95/13.01/126 & 22.30/17.55/13.82/130 & 20.61/15.39/12.42/120 & 43.44/31.56/29.50/184 & 43.86/32.27/29.98/183 & 42.12/31.05/29.04/182  \\
	\cline{2-3}
	& \multirow{2}{*}{1}  & Handcrafted CA & 22.93/17.02/14.04/129 & 23.49/18.26/14.81/132 & 22.30/16.72/13.87/123 & 44.08/32.03/30.03/182 & 43.70/32.24/30.03/182 & 41.93/\textbf{31.28}/29.21/183  \\
	 & & L-Global CA & 23.54/17.90/14.71/131 & 23.40/18.19/14.72/132 & 23.29/17.39/14.95/125 & 43.45/32.79/30.52/182 & 43.95/33.56/31.06/185 & 42.02/\textbf{31.28}/29.21/183  \\
                               & & S-Global CA  & 25.84/18.66/16.23/139 & 24.35/18.83/15.44/134 & 25.58/19.04/16.36/139 & 44.20/33.00/30.73/186 & 44.25/34.16/31.34/186 & 43.07/31.60/29.45/187   \\
                               & & L-Classwise CA & \textbf{26.05}/\textbf{18.68}/\textbf{16.14}/\textbf{140} & \textbf{24.50}/\textbf{18.81}/\textbf{15.68}/\textbf{135} & \textbf{24.44}/\textbf{18.62}/\textbf{15.70}/\textbf{137} & \textbf{44.44}/\textbf{33.45}/\textbf{31.21}/\textbf{187} & \textbf{44.23}/\textbf{33.69}/\textbf{31.16}/\textbf{187} & \textbf{42.71}/31.18/\textbf{29.29}/\textbf{184}  \\
                              
	\cline{2-3}
	& \multirow{2}{*}{3} & Handcrafted CA & 22.87/17.64/14.42/127 & 23.11/18.70/15.11/131 & 23.00/16.99/14.19/126 & \textbf{43.87}/33.03/30.47/\textbf{184} & 43.31/32.02/29.76/182 & \textbf{42.29}/31.03/29.01/\textbf{185}  \\
                               & & L-Global CA & 25.12/\textbf{19.55}/\textbf{16.39}/136 & 23.72/19.55/15.71/136 & 24.98/18.51/15.42/135 & 43.45/34.10/31.07/183 & 43.22/32.97/30.47/182 & 41.25/32.45/29.25/182  \\
              & & S-Global CA & 25.02/19.25/16.12/137 & 24.82/19.66/16.16/140 & 25.37/18.42/15.58/135 & 44.21/33.08/30.52/185 & 43.08/34.04/30.95/184 & 43.00/31.41/29.48/185 \\
                               & & L-Classwise CA  & \textbf{25.27}/19.26/16.36/\textbf{137} & \textbf{25.30}/\textbf{20.02}/\textbf{16.79}/\textbf{141} & \textbf{25.46}/\textbf{18.64}/\textbf{15.74}/\textbf{136} & 43.29/\textbf{34.81}/\textbf{31.25}/183 & \textbf{43.82}/\textbf{33.51}/\textbf{30.95}/\textbf{184} & 41.28/\textbf{32.61}/\textbf{29.26}/182  \\

	\hline
	\end{tabular} 
	}
	\vspace{0.1cm}
	\caption{\textcolor{black}{The performance (in $\%$) of different deep context networks on Corel5k. A quadruplet $\cdot/\cdot/\cdot/\cdot$ stands for $\mathbf{R}/\mathbf{P}/\mathbf{F}/\mathbf{N}_{+}$. In these experiments, $r$ corresponds to the radius of the disk that supports the context.} \label{tab:corelresults}}
\end{table*}

\subsection{Corel5k benchmark}
Corel5k~\cite{Duygulu2002} is another widely-used benchmark for image annotation which consists in 4,999 annotated images with a vocabulary of up to 200 keywords. This dataset is split into two subsets: 4,500 images for training and the rest for testing. Following the standard protocol in~\cite{Duygulu2002}, each test image is annotated with up to 5 keywords and performances are measured using the mean precision and recall over keywords (denoted as $\mathbf{P}$ and $\mathbf{R}$ respectively) as well as the F-scores (referred to as $\mathbf{F}$) and the number of keywords with non-zero recall (denoted as $\mathbf{N}_{+}$); again, higher values of these measures imply better performances. 

As in ImageCLEF, we also rescale images to the median dimension of $400\times500$ pixels, and we partition each image into a regular grid of $2\times2$ and $4\times5$ cells; note that grids of $8\times10$ cells are not investigated in our experiments as the underlying performances are clearly inferior (as shown in Tab.~\ref{tab:imageclefresults}). Each cell is again described with handcrafted features (namely BoWs on a code-book of $500$ words evaluated on dense SIFTs) and also deep pre-trained VGG features. We also consider different kernel maps initializations including linear, polynomial and histogram intersection as already discussed  for ImageCLEF. Since categories are highly unbalanced in Corel5k, we learn ensembles of binary SVMs; for each concept, ten SVMs are trained on all the positive data (belonging to that concept) and a random subset (from the remaining negative data) whose cardinality is three times larger than the positive set. The decision score on a given test image, w.r.t a given concept, is taken as the average score of the ten underlying SVMs. 

Tab.~\ref{tab:corelresults} shows performances for different settings including context-free, handcrafted and learned (stationary and classwise) context-aware kernel networks for different regular grids. For most of the settings, we observe a clear gain of different context-nets when trained on top of BoW features; indeed, the gain in $\mathbf{R}$/$\mathbf{P}$/$\mathbf{F}$/$\mathbf{N}_{+}$ reaches $1.0$/$0.7$/$1.0$/$2$ points for learned contexts, $2.1$/$1.9$/$1.8$/$14$ for classwise contexts and $3.3$/$2.3$/$2.5$/$16$ for stationary ones, all obtained using histogram intersection initial map and a grid of $4\times5$ cells with $r=1$. We also observe a clear gain when using VGG features; this gain reaches $1.0$/$0.8$/$0.8$/$3$ points for  learned context,  $1.6$/$1.4$/$1.2$/$6$ for classwise one and $1$/$8$/$1.4$/$1.2$/$8$ for stationary context, all obtained using polynomial initial map and a grid of $2\times2$ cells with $r=1$. It is worth noticing that the gain of classwise context is not always consistent due to the large number of training parameters (w.r.t the size of training data) compared to stationary context which is relatively less subject to overfitting as its parameters are shared.\\  

\indent Tab.~\ref{tab:corelcomparisonother} shows a comparison of the proposed approach against the related work. These comparative methods (namely LDMKL~\cite{Jiutip2017}, wTKML~\cite{Vo2012}, TagPop $\sigma$ML~\cite{Guillaumin2009}) rely on a battery of handcrafted features (GIST, 8 types of BoWs, etc.) and on  context modeling (using $k$-NN in TagPop $\sigma$ML~\cite{Guillaumin2009}, 2PKNN-ML~\cite{Verma2012} and Laplacian operators in DMKL~\cite{Jiutip2017}). In contrast, our method  --- in spite of using a single BoW --- is still competitive; this is essentially due to the discrimination power of the learned contexts which catch-up with these extensively-tuned handcrafted techniques. Further comparisons involving deep features show a clearer trend and a better improvement against other methods including CNN~\cite{Murthyicmr2015}, ResNet~\cite{HeZhangCVPR2016}, DKN~\cite{Jiutip2017} and DMN~\cite{JiuPR2019} as well as LNR+2PKNN~\cite{Zhang2018}. Note that the latter  models bi-modal ``image and textual'' contexts using listwise neural ranking and nearest neighbor search. We believe that adding extra modalities (mainly textual information) could bring an extra-gain to our context learning; this issue, out of the main scope of this paper, will be addressed as a future work. \\

\noindent Finally, Fig.~(\ref{fig:contextevolution}, bottom) displays the learned layerwise context using histogram intersection as initial map  and a grid of $4\times5$ cells with $r=1$. Fig.~(\ref{fig:contextexamples}, third and fourth rows) displays handcrafted  vs. learned (layerwise vs. stationary  and global vs. classwise) contexts (in the second layer of the underlying networks) superimposed on images of Corel5k using a grid of $4\times5$ cells with $r=1$. The second row in Fig.~\ref{fig:annotationexamples} shows image instances and their annotation results respectively using context-free,  handcrafted vs. learned context networks.

\begin{table}[htbp]
    \centering
\resizebox{0.5\textwidth}{!}{
    \begin{tabular}{c|c|c|ccc}
    \hline
      Method & Learned & context & $\mathbf{R}$ & $\mathbf{P}$ & $\mathbf{N_{+}}$ \\
      & Input feat. & & & &  \\
      \hline
      CRM~\cite{Lavrenko2003} & no & no & 19 & 16 & 107 \\
      InfNet~\cite{Metzlercivr04} & no & no & 24 & 17 & 112 \\ 
      JEC-15~\cite{Makadia2008} & no & yes & 33 & 28 & 140 \\
      FT DMN+SVM~\cite{JiuPR2019}  & no & no & 35 & 21 & 168 \\
      3-layer DKN+SVM~\cite{Jiutip2017}& no & no & 38 & 26 & 158 \\
      TagPop $\sigma$ML~\cite{Guillaumin2009} & no & yes & 42 & 33 & 160 \\
      wTKML~\cite{Vo2012}&  no & yes & 42 & 21 & 173 \\
      KSVM-VT~\cite{Verma2013} & no & no & 42 & 32 & 179 \\
      LDMKL~\cite{Jiutip2017} & no & yes & 44 & 29 & 179 \\
      2PKNN-ML~\cite{Verma2012} & no & yes & 46 & 44 & 191 \\
      L-Global CA (proposed)  & no & yes & 25 & 19 & 137 \\
      S-Global CA (proposed) & no & yes & 26 & 20 & 141  \\
      L-Classwise CA (proposed) & no & yes & 26 & 20  & 142 \\
      \hline
      ResNet~\cite{HeZhangCVPR2016} + SVM & yes & no & 35 & 22 & 161 \\
      FT DMN+SVM~\cite{JiuPR2019}  & yes & no & 38 & 23 & 169 \\ 
      CNN-R~\cite{Murthyicmr2015} &yes & yes &  41 & 32 & 166 \\
      3-layer DKN+SVM~\cite{Jiutip2017} & yes & no & 43 & 25 & 180 \\
      LNR+2PKNN~\cite{Zhang2018} & yes & yes & 52 & 43 & 192 \\
      L-Global CA (proposed)  & yes & yes & 45 & 36 & 189 \\
      S-Global CA (proposed) & yes & yes & 46 & 36 &  194 \\
      L-Classwise CA (proposed) & yes & yes & 45 & 36 & 192 \\
       
        \hline
    \end{tabular}
    }
    \vspace{0.1cm}
    \caption{Extra comparison of the proposed deep context-aware kernel networks w.r.t. the related work. In this table, FT stands for Fine-Tuned. Results corresponding to BoW features (referred to as ``proposed'' in the first part of the table) are obtained using (i) the polynomial kernel map on a grid of $2\times2$ cells with $r=1$ for Global contexts and (ii) the HI kernel map on a grid of $2\times2$ cells with $r=1$ for Classwise contexts. Results corresponding to deep VGG features (referred to as ``proposed'' in the second part of the table) are obtained using the polynomial kernel map on a grid of $2\times2$ cells with $r=1$ for all the learned context variants.\label{tab:corelcomparisonother}}
  \end{table}

\begin{figure*}[tbp]
\begin{center}
\includegraphics[scale=0.35]{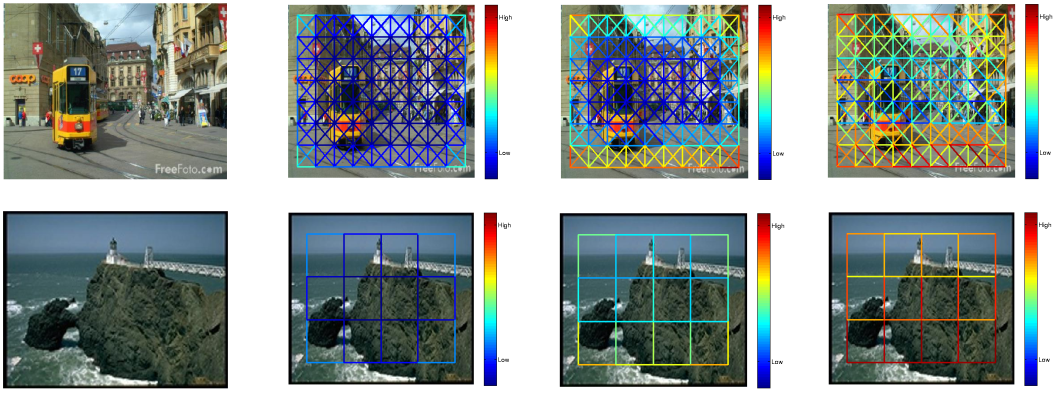}
\end{center}
\caption{\textcolor{black}{This figure shows different context variants on two examples taken from ImageCLEF (top) and Corel5k (bottom); original images (first column), handcrafted contexts (second column), global learned context in the first layer (third column), global learned context in the second layer (fourth column). These results are obtained using a linear kernel map initialization and VGG features with $r=1$  and a grid of $8\times10$ and $4\times5$ cells respectively on the two datasets. Handcrafted context matrices are obtained by normalizing each row (cell) in these matrices by the number of its spatial neighbors, that's why the cells in the four corners have larger values. In all these results, the importance of the context of a given cell is shown with colored connections to its neighbors using a particular color-map; warmer colors (close to red) correspond to important relationships while the cooler ones are less important  (better to zoom the PDF version). } \label{fig:contextevolution}}
\end{figure*}

\begin{figure*}[tbp]
\begin{center}
\includegraphics[scale=0.35]{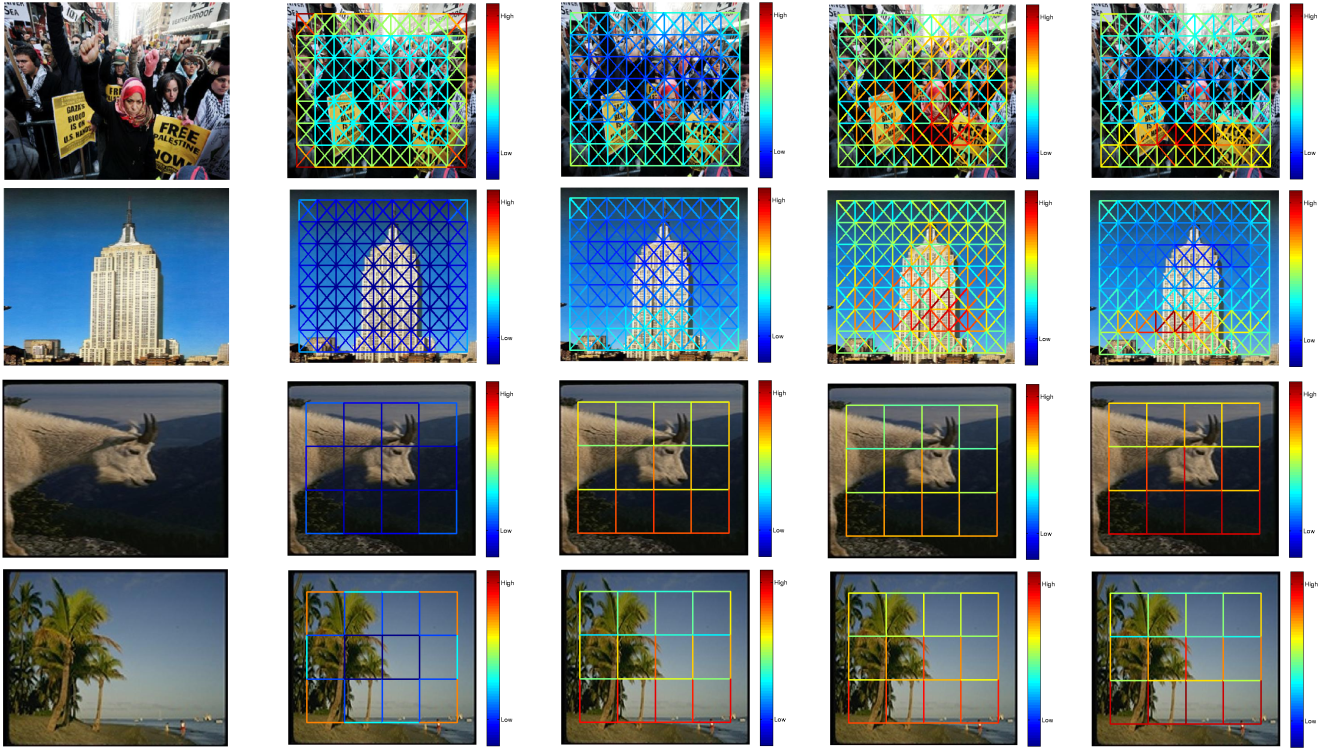}
\end{center}
\caption{\textcolor{black}{This figure shows original images (first column), handcrafted (second column), learned global (third column),  learned stationary (fourth column) and learned classwise contexts (fifth column) using VGG features with $r=1$ and grids of $8\times10$ and $4\times5$ cells respectively for ImageCLEF and Corel5k databases; HI and linear kernel maps are respectively used on these two datasets. For stationary context, it is clear that the importance of the underlying prominent area is strengthened compared to layerwise context and for classwise context, the contribution of background is weakened while the underlying prominent area boosted.}
\textcolor{black}{For  instance, regarding the  concepts ``cityscape'' and "mountain"  (shown in the second and third rows respectively), it is clear that the areas around these concepts are  strengthened compared to the other areas in these scenes.} (better to zoom the PDF version).} \label{fig:contextexamples}
\end{figure*}

\begin{figure*}[tbp]
\begin{center}
\includegraphics[angle=0,width=1\linewidth]{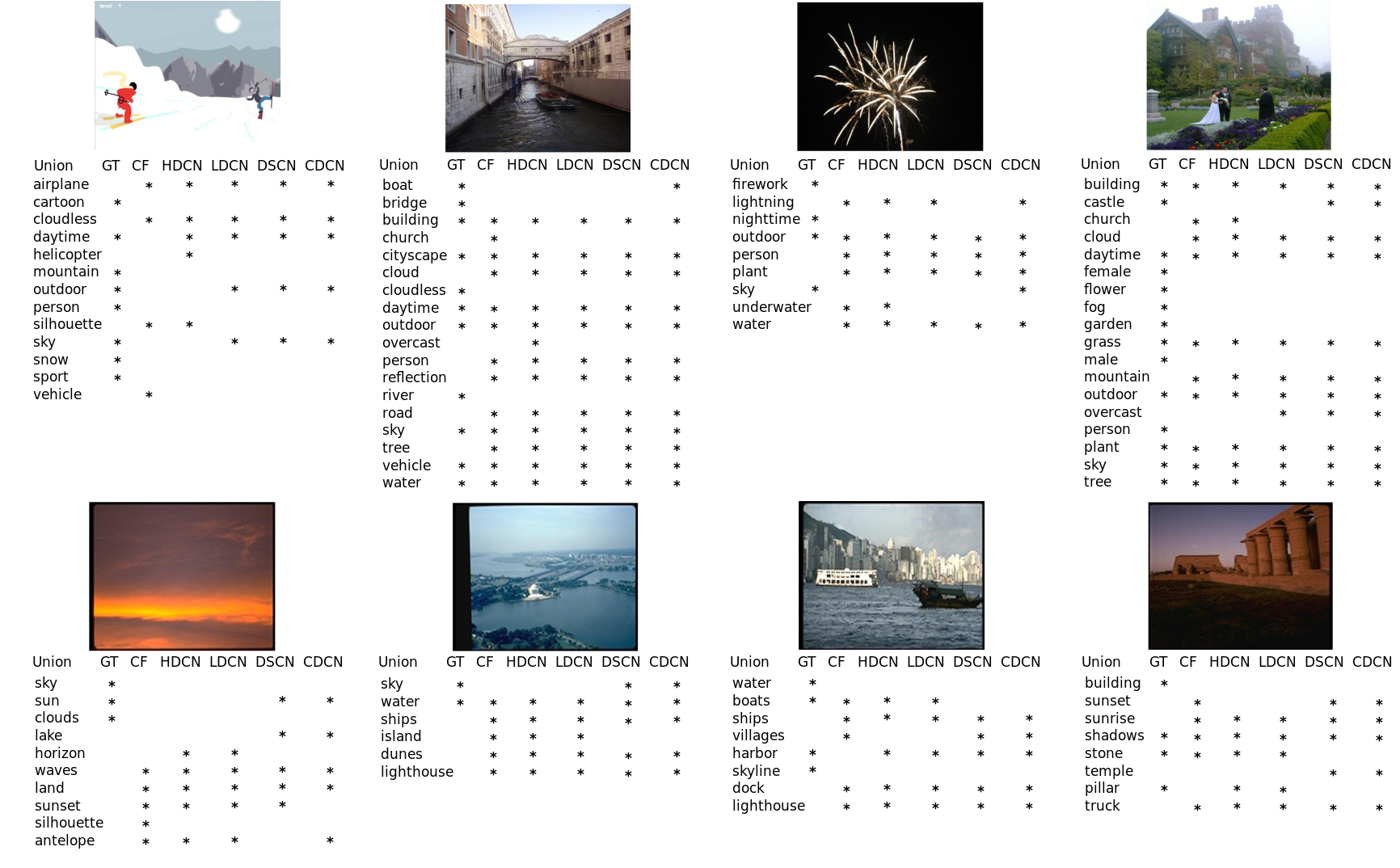}
\end{center}
\caption{\textcolor{black}{These figures show some annotation examples using (from left-to-right on each image): context-free kernels (``CF''),  handcrafted and learned (layerwise, stationary and classwise) context-aware kernel networks. ``GT'' stands for ground-truth annotations and the stars refer to the presence of a given concept in a test image. Results shown in the first row correspond to ImageCLEF while those in the second row to Corel5k. All these results are obtained using polynomial kernel map initialization and VGG features on a grid of $2\times2$ cells with $r=1$.} \label{fig:annotationexamples}}
\end{figure*}

\section{Conclusion}\label{concl}

In this paper we introduce a novel deep context-aware kernel network that considers context learning as as part of kernel design. The proposed method is based on a particular deep network architecture whose parameters --- trained ``end-to-end'' ---  model the contextual relationships between visual patterns (cells) into images. Different variants of contexts are investigated including layerwise, stationary and classwise. While stationary contexts allow us to reduce overfitting, classwise ones make it possible to further enhance the performances by making context class-dependent. 
Extensive experiments conducted on the challenging ImageCLEF and Corel5k benchmarks, show a clear and a consistent gain of classifiers trained on top of the learned context networks w.r.t. classifiers trained using handcrafted context networks as well as context-free ones. As a future work, we are currently investigating the issues of (i) the integration of attention mechanisms into our context networks in order to model primitive  saliency in images  and (ii) the use of a priori knowledge (mainly structures) in context learning; we believe that these two extensions will further enhance the performances. 

\section*{Acknowledgment}
This work was supported by a grant from the National Natural Science Foundation of China (No.~61806180), by a grant from Key Scientific Research Projects of Higher Education of He'nan Province of China (No.~19A520037) and also partially by a grant from the research agency ANR (Agence Nationale de la Recherche) of France under the MLVIS project (ANR-11-BS02-0017).

\end{document}